\documentclass{article}

\usepackage{ijcai07}
\usepackage[latin1]{inputenc}

\usepackage{xspace}
\usepackage{amsmath}
\usepackage{amssymb}
\usepackage{amscd}
\usepackage{pifont}
\usepackage{psfrag}
\usepackage{qtree}
\usepackage{epsfig}
\usepackage{amscd}
\usepackage{enumerate}

\def\Reels{\hbox{$\rm I\!R$}} % realise par DBR

\def\na#1{\textsc{#1}\xspace}
\def\fm#1{\text{\texttt{#1}}} % use ams

\def\aca{\na{aka}}

\def\ecbd{\na{kdd}}
\def\cabamaka{\na{Ca\-ba\-ma\-kA}}

\def\charm{\na{Charm}}

\def\conceptA{\fm{A}}
\def\conceptB{\fm{B}}
\let\conceptC=\concept
\def\conceptD{\fm{D}}

\def\conceptP{\fm{P}}
\def\conceptQ{\fm{Q}}

\def\range(#1, #2){#2\fm{.}#1}
\def\interpretation{{\mathcal{I}}}
\def\interp{^{\interpretation}}
\def\domaine#1{\Delta_{#1}}
\def\domaineInterpretation{\domaine{\interpretation}}
\def\fonctionInterpretation{\cdot\interp}
\def\subsumepar{\sqsubseteq}

\def\baseDeConnaissances{\fm{KB}}

\def\ET{\mathrel{\sqcap}}
\def\fa#1{\fm{#1}} % fonte pour les attributs
\def\={\mathrel{\text{\textbf{:}}}}
\def\intEnt[#1, #2]{[#1, #2]}
\def\intEntOG]#1, #2]{~]#1, #2]} % OG = ouvert à gauche
\def\intEntOD[#1, #2[{[#1, #2[~} % OD = ouvert à droite
\def\intEntO]#1, #2[{~]#1, #2[} % O = ouvert à gauche et à droite
\def\intRee[#1, #2]{[#1 ; #2]}
\def\intReeOD[#1, #2[{[#1 ; #2[~}
\def\intReeOG]#1, #2]{~]#1 ; #2]}
\def\omega{\top}
\def\concretR{\mathrm{\textsf{R}}}

\def\interpR{^{\concretR}}
\def\Ontologie{{\mathcal{O}}}
\def\DecouleOntologie{\vDash_{\Ontologie}}
\def\af{\fm{r}} % role !
\def\cf{\fm{g}}
\def\contrainte{\fm{c}}
\def\contrainteD{\fm{d}}
\def\Contraintes#1{\fm{Cstraints}_{#1}}
\def\SUPE#1{\geq_{#1}}

\def\INF#1{<_{#1}}
\def\Patient{\fm{Patient}}

\def\SeinGauche{\fm{Left-Breast}}
\def\Sein{\fm{Breast}}

\def\Curage{\fm{Curettage}}
\def\Mastectomie{\fm{Mastectomy}}
\def\MastectomieTotale{\fm{Radical-Mastectomy}}
\def\MastectomiePartielle{\fm{Partial-Mastectomy}}

\def\age{\fa{age}}

\def\tumeur{\fa{tumor}}   %% anglais (GB)

\def\taille{\fa{size}}
\def\localisation{\fa{localization}}

\def\rapc{\na{cbr}}

\def\cible{\fm{tgt}}
\def\source{\fm{srce}}
\def\sol(#1){\fm{Sol}(#1)}

\def\prb{\fm{pb}}

\def\solut{\fm{sol}}
\def\src#1{\source_{#1}}
\def\cassource{\fm{srce-case}}
\def\cassourceN#1{\cassource_{#1}}
\def\Problemes{\fm{Problems}}
\def\Solutions{\fm{Solutions}}
\def\basedecas{\fm{CB}}
\def\basedecasdadaptation{\basedecas_{\fm{A}}}

\def\diff#1{\Delta#1}
\def\diffPb{\diff{\prb}}
\def\diffSol{\diff{\solut}}

\def\CA{\fm{AK}}
\def\Proprietes{{\mathcal{P}}}
\def\support{\fm{support}}

\def\form{\Phi} % formatage
\def\Eg{^{\fm{=}}}
\def\Pb{_{\prb}}
\def\Sl{_{\solut}}
\def\Moins{^{\fm{-}}}
\def\Plus{^{\fm{+}}}
\def\cm{\,\text{cm}}
\def\dans{\rightarrow}

\newtheorem{remarqueital}{Remarque}
  {\begin{remarqueital}\bgroup\rm}%
  {\egroup\end{remarqueital}} % \ \\[-6mm]}
\def\fm#1{\text{\texttt{#1}}}
\def\dans{\longrightarrow}
 {\bgroup
  \par\noindent\textbf{Exemple : }
  \em}
 {\egroup\par}

\def\num#1{$#1\degre\!/$\xspace}
\def\num#1{\strut}

\def\etape#1{\ensuremath{(s_{#1})}}

\def\dec{\fm{dec}}
\def\decN#1{\dec_{#1}}

\def\moins{\mathrel{\backslash}}

\def\conceptA{\fm{A}}
\def\conceptD{\fm{D}}

\def\subsumepar{\sqsubseteq}

\def\ET{\,\mathrel{\sqcap}\,}

\def\age{\fm{age}}

\def\tumeur{\fm{tumor}}
\def\localisation{\fm{localization}}

\def\transaction{T}
\def\motif{I}
\def\motifJ{J}
\def\seuil{\sigma}

\usepackage{ifthen}
\newcounter{nbLignes}
\newcounter{nbLignesMoinsUn}

\title{Case Base Mining for Adaptation Knowledge Acquisition}
\author{M. d'Aquin$^{\text{1, 2}}$,
       F. Badra$^{\text{1}}$,
       S. Lafrogne$^{\text{1}}$, \\
       \bf
       J. Lieber$^{\text{1}}$,
       A. Napoli$^{\text{1}}$,
       L. Szathmary$^{\text{1}}$ \\
       $^{\text{1}}$ LORIA (CNRS, INRIA, Nancy Universities) BP 239, 54506 Vand{\oe}uvre-l{{\`e}s}-Nancy, France, \\
       \texttt{\{daquin, badra, lafrogne, lieber, napoli, szathmar\}@loria.fr}
       \\
       $^\text{2}$ Knowledge Media Institute (KMi),
       the Open University, Milton Keynes, United Kingdom, \\
       \texttt{m.daquin@open.ac.uk}}

\begin{document}
%%%%%%%%%%%%%%%%

\maketitle

\bibliographystyle{named}

\begin{abstract}
In case-based reasoning, the adaptation of a source case in order to
 solve the target problem is at the same time crucial and difficult
 to implement.
The reason for this difficulty is that, in general, adaptation strongly
 depends on domain-dependent knowledge.
This fact motivates research on adaptation knowledge acquisition
 (\aca).
This paper presents an approach to \aca based on the principles and
 techniques of knowledge discovery from databases and data-mining.
It is implemented in \cabamaka,
 a system that explores the variations within
 the case base to elicit adaptation knowledge.
This system has been successfully tested in an application of
 case-based reasoning to decision support in the domain of breast
 cancer treatment.
\end{abstract}

%        ------------
\section{Introduction}
%        ------------
\label{sec:introduction}

Case-based reasoning (\rapc~\cite{Riesbeck-Schank89})
 aims at solving a target problem thanks
 to a case base.
A case represents a previously solved problem
 and may be seen as a pair $(\text{problem}, \text{solution})$.
A \rapc system selects a case from the case base
 and then
 adapts the associated solution, requiring domain-dependent knowledge
 for adaptation.
The goal of adaptation knowledge acquisition (\aca)
 is to
 detect and extract
 this knowledge.
This is the function of the semi-automatic system
 \cabamaka,
 which applies
 principles of knowledge discovery from databases (\ecbd)
 to \aca,
 in particular frequent itemset extraction.
This paper presents the system \cabamaka:
 its principles, its implementation and
 an example of adaptation rule discovered
 in the framework of an application
 to breast cancer treatment.
The originality of \cabamaka lies essentially in the approach of \aca
 that uses a powerful learning technique that is guided by a domain
 expert,
 according to the spirit of \ecbd.
This paper proposes an original and working approach to \aca,
 based on \ecbd techniques.
In addition, the \ecbd process is performed on a knowledge base
 itself, leading to the extraction of \emph{meta-knowledge},
 i.e. knowledge units for manipulating other knowledge units.
This is also one of the rare papers trying to build an effective
 bridge between knowledge discovery and case-based reasoning.

The paper is organized as follows.
Section~\ref{sec:rapc-adaptation} presents basic notions
 about \rapc and adaptation.
Section~\ref{sec:travaux-aca} summarizes researches on \aca.
Section~\ref{sec:cabamaka} describes the system \cabamaka:
 its main principles, its implementation and examples of
 adaptation knowledge acquired from it.
Finally, section~\ref{sec:conclusion} draws some conclusions
 and points out future work.

%        --------------------
\section{\rapc and Adaptation}
%        --------------------
\label{sec:rapc-adaptation}

A case in a given \rapc application encodes a problem-solving episode
 that is represented by a problem statement $\prb$ and an associated
 solution $\sol(\prb)$.
The case is denoted by the pair $(\prb, \sol(\prb))$ in
 the following.
Let $\Problemes$ and $\Solutions$
 be the set of problems and the set of solutions of the application
 domain,
 and ``is a solution of'' be a binary relation on
 $\Problemes\times\Solutions$.
In general, this relation
 is not known in the whole
 but at least a finite
 number of its instances
 $(\prb, \sol(\prb))$ is known and constitutes the case base
 $\basedecas$.
An element of $\basedecas$ is called a \emph{source case}
 and is denoted by $\cassource=(\source, \sol(\source))$,
 where $\source$ is a \emph{source problem}.
In a particular \rapc session, the problem to be solved
 is called \emph{target problem}, denoted by $\cible$.

A case-based inference associates to $\cible$ a solution $\sol(\cible)$,
 with respect to
 the case base $\basedecas$ and to
 additional knowledge bases, in particular $\Ontologie$, the
 \emph{domain ontology}
 (also known as domain theory or domain knowledge)
 that usually introduces the concepts and terms
 used to represent the cases.
It can be noticed that the research work presented in
 this paper is based on the
 assumption that there exists a domain ontology associated with
 the case base,
 in the spirit of knowledge-intensive \rapc~\cite{aamodt90}.

A classical decomposition of \rapc relies on
 the steps of retrieval and adaptation.
\emph{Retrieval} selects
 $(\source, \sol(\source))\in\basedecas$ such that $\source$
 is similar to $\cible$ according to some similarity
 criterion.
The goal of adaptation is to solve $\cible$ by modifying
 $\sol(\source)$ accordingly.
Thus, the profile of the adaptation function is
\begin{equation*}
  \fm{Adaptation} :
  ((\source, \sol(\source)), \cible)
  \mapsto
  \sol(\cible)
\end{equation*}

The work presented hereafter is based on the following model of
 adaptation, similar to
 \emph{transformational analogy}~\cite{carbonell83}:
\begin{dingautolist}{192}
\item\label{etape:appariement}
  $(\source, \cible)\mapsto\diffPb$,
   where $\diffPb$ encodes the similarities and dissimilarities
   of the problems $\source$ and $\cible$.
\item\label{etape:parallele}
  $(\diffPb, \CA)\mapsto\diffSol$,
  where $\CA$ is the adaptation knowledge
  and
  where $\diffSol$ encodes the similarities and dissimilarities of
  $\sol(\source)$ and the forthcoming $\sol(\cible)$.
\item\label{etape:modification}
  $(\sol(\source), \diffSol)\mapsto\sol(\cible)$,
  $\sol(\source)$ is modified into $\sol(\cible)$ according
   to $\diffSol$.
\end{dingautolist}

Adaptation is generally supposed to be
 domain-dependent in the sense that it relies on
 domain-specific adaptation knowledge.
Therefore, this knowledge has to be acquired.
This is the purpose of \emph{adaptation knowledge acquisition}
 (\aca).

%        --------------------
\section{Related Work in \aca}
%        --------------------
\label{sec:travaux-aca}

The notion of adaptation case is introduced
 in~\cite{leake96b}.
The system \textsc{dial} is a case-based planner in the domain of
 disaster response planning.
Disaster response planning is the initial strategic planning used to
 determine how to assess damage, evacuate victims, etc. in response to
 natural and man-made disasters such as earthquakes and chemical
 spills.
To adapt a case, the \textsc{dial} system performs either a case-based
adaptation or a rule-based adaptation.
The case-based adaptation attempts to retrieve an adaptation case
describing the successful adaptation of a similar previous adaptation
problem.
An adaptation case represents an adaptation as the combination of
transformations (e.g. addition, deletion, substitution) plus memory
search for the knowledge needed to operationalize the transformation
(e.g. to find what to add or substitute), thus reifying the principle:
\emph{adaptation = transformations + memory search}.
An adaptation case in \textsc{dial} packages information about the
 context of an adaptation, the derivation of its solution, and the
 effort involved in the derivation process.
The context information includes characteristics of the problem for
which adaptation was generated, such as the type of problem, the value
being adapted, and the roles that value fills in the response plan.
The derivation records the operations needed to find appropriate
values in memory, e.g. operations to extract role-fillers or other
information to guide the memory search process.
Finally, the effort records the actual effort expended to find the
solution path.
It can be noticed that the core idea of ``transformation'' is also
 present in our own adaptation knowledge extraction.

In~\cite{jarmulak01}, an approach to \aca is presented
 that produces a set of \emph{adaptation cases},
 where an adaptation case is the representation
 of a particular adaptation process.
The adaptation case base, $\basedecasdadaptation$,
 is then used for further adaptation steps:
 an adaptation step itself is based on \rapc,
 reusing the adaptation cases of $\basedecasdadaptation$.
$\basedecasdadaptation$ is built as follows.
For each $(\src1, \sol(\src1))\in\basedecas$,
 the retrieval step of the \rapc system using the case base
 $\basedecas$ without $(\src1, \sol(\src1))$
 returns a case $(\src2, \sol(\src2))$.
Then, an adaptation case is built based on both source cases
 and is added to $\basedecasdadaptation$.
This adaptation case encodes $\src1$, $\sol(\src1)$,
 the difference between $\src1$ and $\src2$
 ($\diffPb$, with the notations of this paper)
 and the difference between $\sol(\src1)$ and $\sol(\src2)$
 ($\diffSol$).
This approach to \aca and \rapc has been successfully tested
 for an application to the design of tablet formulation.

The idea of the research presented in~\cite{hanney96,hanney97}
 is to exploit the variations between source cases to learn
 adaptation rules.
These rules compute variations on solutions from variations
 on problems.
More precisely, ordered pairs $(\cassourceN1, \cassourceN2)$
 of similar source cases are formed.
Then, for each of these pairs,
 the variations between the problems $\src1$ and $\src2$
 and the solutions $\sol(\src1)$ and $\sol(\src2)$
 are represented
 ($\diffPb$ and $\diffSol$).
Finally, the adaptation rules are learned, using as training
 set the set of the input-output pairs $(\diffPb, \diffSol)$.
This approach has been tested in two domains:
 the estimation of the price of flats and houses,
 and the prediction of the rise time of a servo mechanism.
The experiments have shown that the \rapc system using
 the adaptation knowledge acquired from the automatic system
 of \aca
 shows a better performance compared to the \rapc system
 working without adaptation.
This research has influenced our work
 that is globally based on similar ideas.

\cite{shiu01_COURT} proposes a method for case base maintenance
 that reduces the case base to a set of representative cases together
 with a set of general adaptation rules.
These rules handle the perturbation between representative
 cases and the other ones.
They are generated by a fuzzy decision tree algorithm using
 the pairs of similar source cases as a training set.

In~\cite{wiratunga02}, the idea of~\cite{hanney96} is reused
 to extend the approach of~\cite{jarmulak01}:
 some learning algorithms (in particular, C4.5) are applied to the
 adaptation cases of $\basedecasdadaptation$,
 to induce general adaptation knowledge.

These approaches to \aca
 share the idea of exploiting adaptation cases.
For some of them (\cite{jarmulak01,leake96b}),
 the adaptation cases themselves constitute the adaptation
 knowledge (and adaptation is itself a \rapc process).
For the other ones
 (\cite{hanney96,shiu01_COURT,wiratunga02}),
 as for the approach presented in this paper,
 the adaptation cases are the input of a learning process.

%        ---------
\section{\cabamaka}
%        ---------
\label{sec:cabamaka}

We now present the \cabamaka system, for acquiring adaptation
 knowledge.
The \cabamaka system is at present working in the medical
 domain of cancer treatment,
 but it may be reused in other application domains where there
 exist problems to be solved by a \rapc system.

%           ----------
\subsection{Principles}
%           ----------
\label{subsec:principes}

\cabamaka deals with
 \underline{ca}se \underline{ba}se \underline{m}ining for
 \underline{\aca}.
Although the main ideas underlying \cabamaka are shared
 with those presented in~\cite{hanney96},
 the followings are original ones.
The adaptation knowledge that is mined has to be
 validated by experts and has to be associated with explanations
 making it understandable by the user.
In this way, \cabamaka may be considered as a semi-automated
 (or interactive) learning system.
This is a necessary requirement for the medical domain for which
 \cabamaka has been initially designed.

Moreover, the system takes into account every ordered pair
 $(\cassourceN1, \cassourceN2)$
 with $\cassourceN1\neq\cassourceN2$,
 leading to examine $n(n-1)$ pairs of cases for a case base
 $\basedecas$ where $|\basedecas|=n$.
In practice, this number may be rather large since in the present
 application $n\simeq650$ ($n(n-1)\simeq4\cdot10^5$).
This is one reason for choosing for this system efficient \ecbd
 techniques such as \charm~\cite{zaki02AvecMacroCharm}.
This is different from the approach of~\cite{hanney96}, where
 only pairs of \emph{similar} source cases are considered, according
 to a fixed criterion.
In \cabamaka, there is no similarity criterion on which a selection of
 pairs of cases to be compared could be carried out.
 Indeed, the \rapc process in \cabamaka relies on the
 adaptation-guided retrieval principle~\cite{smyth-keane96}, where
 only adaptable cases are retrieved.
Thus, every pair of cases may be of interest, and two cases may appear
 to be similar w.r.t. a given point of view,
 and dissimilar w.r.t. another one.

%          --------------------
\paragraph{Principles of \ecbd.}
%          --------------------
%
The goal of \ecbd is to discover knowledge from databases,
 under the supervision of an analyst (expert of the domain).
A \ecbd session usually relies on three main steps:
 data preparation, data-mining, and interpretation of
 the extracted pieces of information.

\emph{Data preparation} is mainly based on formatting and filtering
 operations.
The formatting operations are used to transform the data into a
 form allowing the application of the chosen data-mining operations.
The filtering operations are used for removing noisy data and
 for focusing the data-mining operation on special subsets of
 objects and/or attributes.

\emph{Data-mining} algorithms are applied for extracting from data
 information units showing some regularities \cite{hand01}.
In the present experiment, the \charm data-mining algorithm that
 efficiently performs the extraction of 
 \emph{frequent closed itemsets}
 (\emph{FCIs}) has been used \cite{zaki02AvecMacroCharm}.
\charm inputs a formal database, i.e. a set of binary
\emph{transactions}, where each transaction $\transaction$
 is a set of binary \emph{items}.
An \emph{itemset} $\motif$ is a set of items, and the support of
 $\motif$, $\support(\motif)$, is the proportion of transactions
 $\transaction$ of the database possessing $\motif$ 
 ($\motif\subseteq\transaction$).
$\motif$ is frequent, with respect to a threshold $\seuil\in[0 ; 1]$,
 whenever $\support(\motif)\geq\seuil$.
$\motif$ is closed if it has no proper superset $\motifJ$
 ($\motif\subsetneq\motifJ$) with the same support.

The \emph{interpretation} step aims at
 interpreting
 the extracted pieces of information,
 i.e. the FCIs in the present case,
 with the help of an analyst.
In this way, the interpretation step produces new knowledge units
 (e.g. rules).

The \cabamaka system relies on these
 main \ecbd steps as explained below.

%          -----------
\paragraph{Formatting.}
%          -----------
%
The formatting step of \cabamaka inputs the case base $\basedecas$
 and outputs a set of transactions
 obtained from the
 pairs $(\cassourceN1, \cassourceN2)$.
It is composed of two substeps.
During the first substep, each
 $\cassource=(\source, \sol(\source))\in\basedecas$ is formatted in
 two sets of boolean properties:
 $\form(\source)$ and $\form(\sol(\source))$.
The computation of $\form(\source)$ consists in translating $\source$
 from the problem representation formalism to $2^{\Proprietes}$,
 $\Proprietes$ being a set of boolean properties.
Some information may be lost during this
 translation,
 for example, when translating a continuous property into
 a set of boolean properties,
 but this loss has to be minimized.
Now, this translation formats an expression $\source$ expressed
 in the framework of the domain ontology $\Ontologie$ to an expression
 $\form(\source)$ that will be manipulated as data,
 i.e. without the use of a reasoning process.
Therefore, in order to minimize the translation loss, it is
 assumed that
\begin{equation}
  \text{if }
  p\in\form(\source)
  \text{ and }
  p\DecouleOntologie{}q
  \quad
  \text{then }
  q\in\form(\source)
  \label{eq:FermetureDeductive}
\end{equation}
 for each $p, q\in\Proprietes$
 (where $p\DecouleOntologie{}q$ stands for
  ``$q$ is a consequence of $p$ in the ontology $\Ontologie$'').
In other words, $\form(\source)$ is assumed to be deductively
 closed given $\Ontologie$ in the set $\Proprietes$.
The same assumption is made for $\form(\sol(\source))$.
How this first substep of formatting is computed in practice depends
  heavily on the representation formalism of the cases and
 is presented, for our application, in
 section~\ref{subsec:implantation}.

The second substep of formatting produces a transaction
 $\transaction=\form((\cassourceN1,\cassourceN2))$
 for each ordered pair of distinct source cases,
 based on the sets of items
 $\form(\src1)$, $\form(\src2)$,
 $\form(\sol(\src1))$ and $\form(\sol(\src2))$.
Following the model of adaptation presented in
 section~\ref{sec:rapc-adaptation}
 (items~\ref{etape:appariement}, \ref{etape:parallele}
  and~\ref{etape:modification}),
 $\transaction$ has to encode the properties of $\diffPb$ and
 $\diffSol$.
$\diffPb$ encodes the similarities and dissimilarities of $\src1$
 and $\src2$, i.e.:
\begin{itemize}
\item
  The properties common to $\src1$ and $\src2$
   (marked by ``$\fm{=}$''),
\item
  The properties of $\src1$ that $\src2$ does not share
   (``$\fm{-}$''), and
\item
  The properties of $\src2$ that $\src1$ does not share
   (``$\fm{+}$'').
\end{itemize}
All these properties are related to problems and thus are marked by
 $\prb$.
$\diffSol$ is computed in a similar way and
 $\form(\transaction)=\diffPb\cup\diffSol$.
For example,
\begin{align}
  &\text{if}
  \quad
  \begin{cases}
    \form(\src1) = \{a, b, c\}
    \quad
    &\form(\sol(\src1)) = \{A, B\} \\
    \form(\src2) = \{b, c, d\}
    \quad
    &\form(\sol(\src2)) = \{B, C\}
  \end{cases}
  \notag\\
  &\text{then}
  \quad
  \transaction = \left\{a\Moins\Pb, b\Eg\Pb, c\Eg\Pb, d\Plus\Pb, A\Moins\Sl, B\Eg\Sl, C\Plus\Sl\right\}
  \label{eq:formatage}
\end{align}
More generally:
\begin{align*}
  \transaction
  &=     \{p\Moins\Pb ~|~ p\in\form(\src1)\backslash\form(\src2)\} \\
  &~\cup \{p\Eg\Pb    ~|~ p\in\form(\src1)\cap\form(\src2)\} \\
  &~\cup \{p\Plus\Pb  ~|~ p\in\form(\src2)\backslash\form(\src1)\} \\
  &~\cup \{p\Moins\Sl ~|~ p\in\form(\sol(\src1))\backslash\form(\sol(\src2))\} \\
  &~\cup \{p\Eg\Sl    ~|~ p\in\form(\sol(\src1))\cap\form(\sol(\src2))\} \\
  &~\cup \{p\Plus\Sl  ~|~ p\in\form(\sol(\src2))\backslash\form(\sol(\src1))\}
\end{align*}

%          ----------
\paragraph{Filtering.}
%          ----------
%
The filtering operations may take place before, between and after the
 formatting substeps, and also after the mining step.
They are guided by the analyst.

%          -------
\paragraph{Mining.}
%          -------
%
The extraction of FCIs is computed thanks to
 \charm (in fact, thanks to a tool based on a \charm-like algorithm)
 from the set of transactions.
A transaction
 $\transaction=\form((\cassourceN1, \cassourceN2))$ encodes a specific adaptation
 $((\src1, \sol(\src1)), \src2)\mapsto\sol(\src2)$.
For example, consider the following FCI:
\begin{equation}
  \motif = \left\{a\Moins\Pb, c\Eg\Pb, d\Plus\Pb, A\Moins\Sl, B\Eg\Sl, C\Plus\Sl\right\}
  \label{eq:exemple-MFF}
\end{equation}
 $\motif$ can be considered as a generalization of a subset of the
 transactions including the transaction $\transaction$
 of equation (\ref{eq:formatage}):
 $\motif\subseteq\transaction$.
The interpretation of this FCI as an adaptation rule is explained
 below.

%          ---------------
\paragraph{Interpretation.}
%          ---------------
%
The interpretation step is supervised by the analyst.
The \cabamaka system provides the analyst with the extracted FCIs
 and facilities for navigating among them.
The analyst may select an FCI, say $\motif$, and interpret $\motif$
 as an adaptation rule.
For example, the FCI
 in equation~(\ref{eq:exemple-MFF})
 may be interpreted in the following terms:
\begin{quote}
  \strut\llap{\textbf{if~}}\strut
  \num1 $a$ is a property of $\source$ but is not a property of $\cible$, \\
  \num2 $c$ is a property of both $\source$ and $\cible$, \\
  \num3 $d$ is not a property of $\source$ but is a property of $\cible$, \\
  \num4 $A$ and $B$ are properties of $\sol(\source)$, and \\
  \num5 $C$ is not a property of $\sol(\source)$ \\
  \strut\hspace{-5mm}{\textbf{then}}\strut
  ~the properties of $\sol(\cible)$ are \\
  \strut~~
  $\form(\sol(\cible)) = (\form(\sol(\source))\moins\{A\})\cup\{C\}$.
\end{quote}
This rule has to be translated from the formalism
 $2^{\Proprietes}$ (sets of boolean properties)
 to the formalism of the adaptation rules of the \rapc system.
The result is an \emph{adaptation rule}, i.e. a rule
 whose left part represents
 conditions on $\source$, $\sol(\source)$ and $\cible$
 and whose right part represents
 a way to compute $\sol(\cible)$.
The role of the analyst is to correct and to validate this adaptation
 rule and to associate an explanation with it.
The analyst is helped in this task by the domain ontology $\Ontologie$
 that is useful for organizing the FCIs and by the already available
 adaptation knowledge that is useful for pruning from the FCIs the
 ones that are
 already known adaptation knowledge.

%           --------------
\subsection{Implementation}
%           --------------
\label{subsec:implantation}

The \cabamaka discovery process relies on the steps described in the
 previous section:
 $\etape1$ input the case base,
 $\etape2$ select a subset of it (or take the whole case base):
           first filtering step,
 $\etape3$ first formatting substep,
 $\etape4$ second filtering step,
 $\etape5$ second formatting substep,
 $\etape6$ third filtering step,
 $\etape7$ data-mining (\charm),
 $\etape8$ last filtering step
 and
 $\etape9$ interpretation.
This process is interactive and iterative:
 the analyst runs each of the $\etape{i}$
 (and can interrupt it),
 and can go back to a previous step
 at each moment.

Among these steps, only the first ones
 ($\etape1$ to $\etape3$)
 and the last one are
 dependent on the representation formalism.
In the following, the step
 $\etape3$ is illustrated in the context of an application.
First, some elements on the application itself and the
 associated knowledge representation formalism are
 introduced.

%          -------------------
\paragraph{Application domain.}
%          -------------------
%
The application domain of the \rapc system we are developing is
 breast cancer treatment:
 in this application, a problem $\prb$ describes a class of
 patients
 with a set of attributes and associated constraints
 (holding on the age of the patient, the size and the localization of
  the tumor, etc.).
A solution $\sol(\prb)$ of $\prb$ is a set of therapeutic decisions
 (in surgery, chemotherapy, radiotherapy, etc.).

Two features of this application must be pointed out.
First, the source cases are \emph{general cases}
 (or \emph{ossified cases} according to the terminology
  of~\cite{Riesbeck-Schank89}):
 a source case corresponds to a class of patients and not to a single
 one.
These source cases are obtained from statistical studies in the cancer
 domain.
Second,
 the requested behavior of the \rapc system
 is to provide a treatment
 and explanations on
 this treatment proposal.
This is why the analyst is required to associate an explanation to a
 discovered adaptation rule.

%          ---------------------------------------------------
\paragraph{Representation of cases and of the domain ontology $\Ontologie$.}
%          ---------------------------------------------------
%
The problems, the solutions, and the domain ontology of the
 application are represented in a light extension of OWL~DL
 (the Web Ontology Language recommended by the
  W3C~\cite{staab04}).
The parts of the underlying description logic that
 are useful for this paper are presented below
 (other elements on description logics, DLs, may be found
  in~\cite{staab04}).

Let us consider the following example:
{%
\def\ns{\!\!\!\!\!}
\def\nsii{\!\!\!\!\!\!}
\begin{equation}
\begin{array}{l l l}
  \source
  \equiv&
  \ns\Patient
  \ET
  &\nsii\exists\age.{\SUPE{45}}
  \ET
  \exists\age.{\INF{70}}
  \\
  &\ns\ET
  \exists\tumeur.(&\nsii\exists\taille.{\SUPE{4}} \\
  &               &\nsii\ET \exists\localisation.
                  \SeinGauche)
\end{array}
  \label{eq:defSource}
\end{equation}
}
$\source$ represents the class of patients
 with an age $a\in[45 ; 70[$,
 and a tumor of size $S\geq4$ centimeters
 localized in the left breast.

The DL representation entities used here are
 atomic and defined
 {concepts} (e.g. $\source$, $\Patient$
 and $\exists\age.{\SUPE{45}}$),
 {roles}
  (e.g. $\tumeur$ and $\localisation$)
 {concrete roles}
 (e.g. $\age$ and $\taille$)
 and constraints
 (e.g. $\SUPE{45}$ and $\INF{70}$).
A \emph{concept} $\conceptC$ is an expression representing a
 class of objects.
A \emph{role}
 $\af$ is a name representing
 a binary relation between objects.
A \emph{concrete role} $\cf$ is a name representing
 a function associating a real number to an object
 (for this simplified presentation, the only concrete
  domain that is considered is $(\Reels, \leq)$,
  the ordered set of real numbers).
A constraint $\contrainte$ represents a subset of
 $\Reels$ denoted by $\contrainte\interpR$.
For example,
 intervals such as
 $\SUPE{45}\interpR=[45;+\infty[$ and
 $\INF{70}\interpR=]-\infty;70[$
 introduce constraints that are used in the application.

A concept is either atomic (a concept name)
 or defined.
A defined concept is an expression of
 the following form:
 $\conceptC\ET\conceptD$,
 $\exists\af.\conceptC$ or
 $\exists\cf.\contrainte$,
 where
 $\conceptC$ and $\conceptD$ are concepts,
 $\af$ is a role,
 $\cf$ is a concrete role and
 $\contrainte$ is a constraint
 (many other constructions exist in the DL, but only
  these three constructions are used here).
Following classical DL presentations~\cite{staab04},
 an \emph{ontology} $\Ontologie$ is a set of axioms,
 where an axiom is a formula of the form
 $\conceptC\subsumepar\conceptD$
 (general concept inclusion)
 or of the form $\conceptC\equiv\conceptD$,
 where $\conceptC$ and $\conceptD$ are two concepts.

The semantics of the DL expressions used hereafter can be
 read as follows.
An interpretation is a pair
 $\interpretation=(\domaineInterpretation, \fonctionInterpretation)$
 where $\domaineInterpretation$ is a non empty set
 (the \emph{interpretation domain})
 and $\fonctionInterpretation$
 is the \emph{interpretation function},
 which maps a concept $\conceptC$ to a set
 $\conceptC\interp\subseteq\domaineInterpretation$,
 a role
 $\af$ to a binary relation
 $\af\interp\subseteq\domaineInterpretation\times\domaineInterpretation$,
 and
 a concrete role $\cf$ to a function
 $\cf\interp:\domaineInterpretation\dans\Reels$.
In the following, all roles $\af$ are assumed to be \emph{functional}:
 $\fonctionInterpretation$ maps $\af$ to a function
 $\af\interp:\domaineInterpretation\dans\domaineInterpretation$.
The interpretation of the defined concepts,
 for an interpretation $\interpretation$, is
 as follows:\linebreak
 $(\conceptC\ET\conceptD)\interp=\conceptC\interp\cap\conceptD\interp$,
 $(\exists\af.\conceptC)\interp$ is the set of
 objects $x\in\domaineInterpretation$ such that
 $\af\interp(x)\in\conceptC\interp$ and
 $(\exists\cf.\contrainte)\interp$ is the set of
 objects $x\in\domaineInterpretation$ such that
 $\cf\interp(x)\in\contrainte\interpR$.
An interpretation $\interpretation$ is a \emph{model}
 of an axiom $\conceptC\subsumepar\conceptD$
 (resp. $\conceptC\equiv\conceptD$) if
 $\conceptC\interp\subseteq\conceptD\interp$
 (resp. $\conceptC\interp=\conceptD\interp$).
$\interpretation$ is a model of an ontology $\Ontologie$
 if it is a model of each axiom of $\Ontologie$.
The inference associated with this representation formalism
 that is used below is the subsumption test:
 given an ontology $\Ontologie$, a concept $\conceptC$ is
 subsumed by a concept $\conceptD$,
 denoted by $\DecouleOntologie\conceptC\subsumepar\conceptD$,
 if for every model $\interpretation$ of $\Ontologie$,
 $\conceptC\interp\subseteq\conceptD\interp$.

More practically, the problems of the \rapc application are
 represented by concepts
 (as
  $\source$
  in (\ref{eq:defSource})).
A therapeutic decision $\dec$ is also represented by a concept.
A solution is a finite set
 $\{\decN1, \decN2, \ldots \decN{k}\}$
 of decisions.
The decisions of the system are represented by atomic concepts.
The knowledge associated with atomic concepts
 (and hence, with therapeutic decisions)
 is given by axioms of the domain ontology $\Ontologie$.
For example, the decision in surgery
 $\dec=\fm{Partial-Mastectomy}$
 represents a partial ablation of the breast:
\begin{align}
  \MastectomiePartielle
  &\subsumepar\Mastectomie
  \notag\\
  \Mastectomie
  &\subsumepar\fm{Surgery}
  \label{eq:axiomesMastectomiePartielle}
  \\
  \fm{Surgery}
  &\subsumepar\fm{Therapeutic-Decision}
  \notag
\end{align}

%          ---------------------------------------------------------
\paragraph{Implementation of the first formatting substep $\etape3$.}
%          ---------------------------------------------------------
%
Both problems and decisions constituting solutions
 are represented by concepts.
Thus, computing $\form(\source)$ and $\form(\sol(\source))$
 amounts to the computation of $\form(\conceptC)$,
 $\conceptC$ being a concept.
A property $p$ is an element of the finite set $\Proprietes$
 (see section~\ref{subsec:principes}).
In the DL formalism, $p$ is represented by a concept $\conceptP$.
A concept $\conceptC$ has the property $p$
 if\linebreak
 $\DecouleOntologie\conceptC\subsumepar\conceptP$.
The set of boolean properties and the set of the corresponding
 concepts are both denoted by $\Proprietes$
 in the following.
Given $\Proprietes$, $\form(\conceptC)$ is simply
 defined as the set of properties $\conceptP\in\Proprietes$ that
 $\conceptC$ has:
\begin{equation}
  \form(\conceptC)
  = \{\conceptP\in\Proprietes
  ~|~
  \DecouleOntologie\conceptC\subsumepar\conceptP\}
  \label{eq:definitionForm1DL}
\end{equation}
As a consequence, if $\conceptP\in\form(\conceptC)$,
 $\conceptQ\in\Proprietes$ and
 $\DecouleOntologie\conceptP\subsumepar\conceptQ$
 then $\conceptQ\in\form(\conceptC)$.
Thus, the implication
 (\ref{eq:FermetureDeductive}) is satisfied.

The algorithm of the first formatting substep
 that has been implemented first
 computes the $\form(\conceptC)$'s for $\conceptC$:
 the source problems and the decisions occurring in their
 solutions,
 and then computes $\Proprietes$ as the union of
 the $\form(\conceptC)$'s.
This algorithm relies on the following set of equations\footnote{%
  This set of equations itself can be seen as a recursive algorithm,
   but is not very efficient since it computes several times the same
   things.
  The implemented algorithm avoids these recalculations by the use of
   a cache.}:
{
\def\ns{\!\!\!\!}
\begin{align*}
  \form(\conceptA)
  &=
  \left\{\conceptB ~\left|~
    \ns\text{
      \begin{tabular}{l}
        $\conceptB$ is an atomic concept \\
	occurring in $\baseDeConnaissances$
        and $\DecouleOntologie\conceptA\subsumepar\conceptB$
      \end{tabular}}\ns\right.\right\}
  \\
  \form(\conceptC\ET\conceptD)
  &=
  \form(\conceptC)\cup\form(\conceptD)
  \\
  \form(\exists\af.\conceptC)
  &=
  \{\exists\af.\conceptP ~|~
  \conceptP\in\form(\conceptC)\}
  \\
  \form(\exists\cf.\contrainte)
  &=
  \left\{\exists\cf.\contrainteD ~\left|~
  \contrainteD\in\Contraintes{\cf}
  \text{ and }
  \contrainte\interpR\subseteq\contrainteD\interpR\right.\right\}
  \\
  \Contraintes{\cf}
  &=\{\contrainte ~|~
  \text{the expression $\exists\cf.\contrainte$ occurs in $\baseDeConnaissances$}\}
\end{align*}
}
 where
 $\conceptA$ is an atomic concept,
 $\conceptC$ and $\conceptD$ are (either atomic or defined) concepts,
 $\af$ is a role,
 $\cf$ is a concrete role,
 $\contrainte$ is a constraint and
 $\baseDeConnaissances$, the knowledge base, is the union of
 the case base and of the domain ontology.

It can be proven that the algorithm for the first formatting substep
 (computing the $\form(\conceptC)$'s and the set of properties
  $\Proprietes$)
 respects (\ref{eq:definitionForm1DL})
 under the following hypotheses.
First, the constructions used in the DL are the ones that have been
 introduced above
 ($\conceptC\ET\conceptD$,
  $\exists\af.\conceptC$ and
  $\exists\cf.\contrainte$, where $\af$ is functional).
Second, no defined concept may strictly subsume an atomic
 concept
 (for every atomic concept $\conceptA$,
  there is no defined concept $\conceptC$ such
  that $\DecouleOntologie\conceptA\subsumepar\conceptC$
  and $\not\DecouleOntologie\conceptA\equiv\conceptC$).
Under these hypotheses, (\ref{eq:definitionForm1DL})
 can be proven by a recursion on the size of $\conceptC$
 (this size is the number of constructions that $\conceptC$
  contains).
These hypotheses hold for our application.
However, an ongoing study aims at finding an algorithm for
 computing the $\form(\conceptC)$'s and $\Proprietes$ in a
 more expressive DL, including in particular negation and
 disjunction of concepts.

For example, let $\source$ be the problem introduced by the
 axiom (\ref{eq:defSource}).
It is assumed that the constraints associated with the
 concrete role
 $\age$ in $\baseDeConnaissances$ are
 $\INF{30}$,
 $\SUPE{30}$,
 $\INF{45}$,
 $\SUPE{45}$,
 $\INF{70}$ and
 $\SUPE{70}$,
 that the constraints associated with the concrete role
 $\taille$ in $\baseDeConnaissances$ are
 $\INF{4}$ and $\SUPE{4}$,
 that there is no concept $\conceptA\neq\Patient$
 in $\baseDeConnaissances$ such that
 $\DecouleOntologie\Patient\subsumepar\conceptA$,
 and
 that the only concept $\conceptA\neq\SeinGauche$ of $\baseDeConnaissances$ such that
 $\DecouleOntologie\SeinGauche\subsumepar\conceptA$ is $\conceptA=\Sein$.
Then, the implemented algorithm returns:
\begin{align*}
\form(\source) = \{&
  \Patient,
  \exists\age.{\SUPE{30}},
  \exists\age.{\SUPE{45}},
  \exists\age.{\INF{70}}, \\
  &
  \exists\tumeur.\exists\taille.{\SUPE{4}},
  \\
  &
  \exists\tumeur.\exists\localisation.\SeinGauche
  \\
  &
  \exists\tumeur.\exists\localisation.\Sein
\}
\end{align*}
And the $7$ elements of $\form(\source)$ are
 added to $\Proprietes$.

Another example, based on the set of axioms
 (\ref{eq:axiomesMastectomiePartielle}) is:
\begin{align*}
\form(\MastectomiePartielle)
   = \{\MastectomiePartielle, \\
   \Mastectomie, \fm{Surgery},
   \fm{Therapeutic-Decision}\}
\end{align*}

%           -------
\subsection{Results}
%           -------
\label{subsec:resultats}

The \cabamaka process piloted by the analyst produces a set of
 FCIs.
With $n=647$ cases and $\sigma=10\%$,
 \cabamaka has given $2344$ FCIs
 in about $2$ minutes (on a current PC).
Only the FCIs with at least a $\fm+$ or a $\fm-$
 in both problem properties and solution properties were kept,
 which corresponds to $208$ FCIs.
Each of these FCIs $\motif$ is presented for interpretation to the
 analyst under a simplified form by removing some of the items that
 can be deduced from the ontology.
In particular
  if $\conceptP\Eg\Pb\in\motif$, $\conceptQ\Eg\Pb\in\motif$
   and $\DecouleOntologie\conceptP\subsumepar\conceptQ$
   then $\conceptQ\Eg\Pb$ is removed from $\motif$.
For example, if $\conceptP=(\exists\age\SUPE{45})\in\Proprietes$,\linebreak
 $\conceptQ=(\exists\age\SUPE{30})\in\Proprietes$
 and $(\exists\age\SUPE{45})\Eg\Pb\in\motif$,
 then, necessarily, $(\exists\age\SUPE{30})\Eg\Pb\in\motif$,
 which is a redundant piece of information.

The following FCI has been extracted from \cabamaka:
\begin{align*}
  \motif =
  \{&
  (\exists\age.\INF{70})\Eg\Pb,
  \\
  &
  (\exists\tumeur.\exists\taille.\INF4)\Moins\Pb,
  (\exists\tumeur.\exists\taille.\SUPE4)\Plus\Pb,
  \\
  &
  \Curage\Eg\Sl,
  \Mastectomie\Eg\Sl,
  \\
  &
  \MastectomiePartielle\Moins\Sl,
  \MastectomieTotale\Plus\Sl
  \}
\end{align*}
It has been interpreted in the following way:
 if $\source$ and $\cible$ both represent classes of patients
 of less than $70$ years old,
 if the difference between $\source$ and $\cible$
 lies in the tumor size of the patients---less than $4\cm$
 for the ones of $\source$ and more than $4\cm$ for the ones of
 $\cible$---and if a partial mastectomy
 and a curettage of the lymph nodes
 are proposed for the $\source$,
 then $\sol(\cible)$ is obtained by substituting
 in $\sol(\source)$
 the partial mastectomy by a radical one.

It must be noticed that
 this example has been chosen for its simplicity:
 other adaptation rules have been extracted that are less easy to
 understand.
More substantial experiments have to be carried out for an effective
 evaluation.

The choice of considering \emph{every} pairs of distinct source cases
 can be discussed.
Another version of \cabamaka has been tested that considers only
 similar source cases, as in~\cite{hanney96}:
 only the pairs
 of source cases such that
 $|\form(\src1)\cap\form(\src2)|\geq{}k$
 were considered
 (experimented with $k=1$ to $k=10$).
The first experiments have not shown yet any
 improvements in the results, compared to the version
 without this constraint ($k=+\infty$),
 and involves the necessity to have the threshold $k$ fixed.

%        --------------------------
\section{Conclusion and Future Work}
%        --------------------------
\label{sec:conclusion}

The \cabamaka system presented in this paper is inspired by
 the research of Kathleen Hanney and Mark T. Keane~\cite{hanney96} and
 by the principles of \ecbd for the purpose of semi-automatic
 adaptation knowledge acquisition.
It reuses an FCI extraction tool developed in our team
 and based on a \charm-like algorithm.
Although
 implemented for a specific application
 to breast cancer treatment decision support, it has been
 designed to be reusable for other \rapc applications:
 only a few modules of \cabamaka are dependent on the formalism
 of the cases and of the domain ontology,
 and this formalism, OWL~DL, is a well-known standard.

One element of future work consists in
 searching for ways of simplifying
 the presentation of the numerous extracted FCIs to the
 analyst.
This involves an organization of these FCIs for the purpose of
 navigation among them.
Such an organization can be a hierarchy of FCIs according to
 their specificities or a clustering of the FCIs in
 themes.

A second piece of future work, still for the purpose of helping the
 analyst, is to study the algebraic structure of all the
 possible adaptation rules associated with the operation
 of composition:
 $r$ is a composition of $r_1$ and $r_2$ if adapting
 $(\source, \sol(\source))$ to solve $\cible$ thanks to $r$
 gives the same solution $\sol(\cible)$ as
 (1) solving a problem $\prb$ by adaptation of
     $(\source, \sol(\source))$ thanks to $r_1$ and
 (2) solving $\cible$ by adaptation of
     $(\prb, \sol(\prb))$ thanks to $r_2$.
The idea is to find a smallest family of adaptation rules,
 $F$, such that the closure of $F$ under composition contains
 the sets of the extracted adaptation rules expressed in the
 form of FCIs.
It is hoped that $F$ is much smaller than $S$ and so requires
 less effort from the analyst while corresponding to the same
 adaptation knowledge.

Another study on \aca for our \rapc system was \aca from experts
 (based on the analysis of the adaptations performed by the experts).
This \aca has led to a few adaptation rules and also to
 \emph{adaptation patterns},
 i.e. general strategies for case-based decision support that
 are associated with explanations but that need to be instantiated to
 become operational.
A third future work is \emph{mixed} \aca,
 that is a combined use of the adaptation patterns and of the
 adaptation rules extracted from \cabamaka:
 the idea is to try to instantiate the former by the latter
 in order to obtain a set of human-understandable and operational
 adaptation rules.

{
%  \bibliography{biblio}

\begin{thebibliography}{}

\bibitem[\protect\citeauthoryear{Aamodt}{1990}]{aamodt90}
A.~Aamodt.
\newblock {Knowledge-Intensive Case-Based Reasoning and Sustained Learning}.
\newblock In L.~C. Aiello, editor, {\em {Proc. of the 9th European Conference
  on Artificial Intelligence (ECAI'90)}}, August 1990.

\bibitem[\protect\citeauthoryear{Carbonell}{1983}]{carbonell83}
J.~G. Carbonell.
\newblock {Learning by analogy: Formulating and generalizing plans from past
  experience}.
\newblock In {R. S. Michalski and J. G. Carbonell and T. M. Mitchell}, editor,
  {\em {Machine Learning, An Artificial Intelligence Approach}}, chapter~5,
  pages 137--161. {Morgan Kaufmann, Inc.}, 1983.

\bibitem[\protect\citeauthoryear{Hand \bgroup \em et al.\egroup
  }{2001}]{hand01}
D.~Hand, H.~Mannila, and P.~Smyth.
\newblock {\em {Principles of Data Mining}}.
\newblock The MIT Press, Cambridge (MA), 2001.

\bibitem[\protect\citeauthoryear{Hanney and Keane}{1996}]{hanney96}
K.~Hanney and M.~T. Keane.
\newblock {Learning Adaptation Rules From a Case-Base}.
\newblock In I.~Smith and B.~Faltings, editors, {\em Advances in Case-Based
  Reasoning -- Third European Workshop, EWCBR'96}, LNAI 1168, pages 179--192.
  Springer Verlag, Berlin, 1996.

\bibitem[\protect\citeauthoryear{Hanney}{1997}]{hanney97}
K.~Hanney.
\newblock {Learning Adaptation Rules from Cases}.
\newblock Master's thesis, {Trinity College, Dublin}, 1997.

\bibitem[\protect\citeauthoryear{Jarmulak \bgroup \em et al.\egroup
  }{2001}]{jarmulak01}
J.~Jarmulak, S.~Craw, and R.~Rowe.
\newblock {Using Case-Base Data to Learn Adaptation Knowledge for Design}.
\newblock In {\em {Proceedings of the 17th International Joint Conference on
  Artificial Intelligence (IJCAI'01)}}, pages 1011--1016. {Morgan Kaufmann,
  Inc.}, 2001.

\bibitem[\protect\citeauthoryear{Leake \bgroup \em et al.\egroup
  }{1996}]{leake96b}
D.~B. Leake, A.~Kinley, and D.~C. Wilson.
\newblock {Acquiring Case Adaptation Knowledge: A Hybrid Approach}.
\newblock In {\em {{AAAI}/{IAAI}}}, volume~1, pages 684--689, 1996.

\bibitem[\protect\citeauthoryear{Riesbeck and Schank}{1989}]{Riesbeck-Schank89}
C.~K. Riesbeck and R.~C. Schank.
\newblock {\em {Inside Case-Based Reasoning}}.
\newblock Lawrence Erlbaum Associates, Inc., Hillsdale, New Jersey, 1989.

\bibitem[\protect\citeauthoryear{Shiu \bgroup \em et al.\egroup
  }{2001}]{shiu01_COURT}
S.~C.~K. Shiu, Daniel~S. Yeung, C.~Hung Sun, and X.~Wang.
\newblock {Transferring Case Knowledge to Adaptation Knowledge: An Approach for
  Case-Base Maintenance}.
\newblock {\em {Computational Intelligence}}, 17(2):295--314, 2001.

\bibitem[\protect\citeauthoryear{Smyth and Keane}{1996}]{smyth-keane96}
B.~Smyth and M.~T. Keane.
\newblock {Using adaptation knowledge to retrieve and adapt design cases}.
\newblock {\em {Knowledge-Based Systems}}, 9(2):127--135, 1996.

\bibitem[\protect\citeauthoryear{Staab and Studer}{2004}]{staab04}
S.~Staab and R.~Studer, editors.
\newblock {\em {Handbook on Ontologies}}.
\newblock Springer, Berlin, 2004.

\bibitem[\protect\citeauthoryear{Wiratunga \bgroup \em et al.\egroup
  }{2002}]{wiratunga02}
N.~Wiratunga, S.~Craw, and R.~Rowe.
\newblock {Learning to Adapt for Case-Based Design}.
\newblock In S.~Craw and A.~Preece, editors, {\em {Proceedings of the 6th
  European Conference on in Case-Based Reasoning (ECCBR-02)}}, LNAI 2416, pages
  421--435, 2002.

\bibitem[\protect\citeauthoryear{Zaki and Hsiao}{2002}]{zaki02AvecMacroCharm}
M.~J. Zaki and C.-J. Hsiao.
\newblock {\charm: An Efficient Algorithm for Closed Itemset Mining}.
\newblock In {\em {SIAM International Conference on Data Mining SDM'02}}, pages
  33--43, Apr 2002.

\end{thebibliography}

}

%%%%%%%%%%%%%%
\end{document}